%% file: kammerer.tex
\documentclass[runningheads]{llncs}
\usepackage[T1]{fontenc}
\usepackage{pgfplots}
\usepackage{amsmath}
\usepackage{bm}
\usepackage[font=small,skip=0pt]{caption}
\usepackage{subcaption}

\begin{document} 

\title{
    Symbolic Regression with Fast Function Extraction and Nonlinear Least Squares Optimization
}
\titlerunning{FFX NLS for Symbolic Regression} 

\author{
    Lukas Kammerer\inst{1,2}\orcidID{0000-0001-8236-4294} \and
    Gabriel Kronberger\inst{1}\orcidID{0000-0002-3012-3189} \and
    Michael Kommenda\inst{1}\orcidID{0000-0003-2049-723X}
}
\authorrunning{Kammerer et al.}

\institute {
    Josef Ressel Center for Symbolic Regression\\
    Heuristic and Evolutionary Algorithms Laboratory\\
    University of Applied Sciences Upper Austria, Hagenberg, Austria\\
    \vspace{0.2cm}
    \and
    Department of Computer Science\\
    Johannes Kepler University, Linz, Austria\\
    \email{lukas.kammerer@fh-hagenberg.at}
}

\maketitle

\renewcommand{\thefootnote}{}
\footnotetext{\hspace{-0em}
Submitted manuscript to be published in \textit{Computer Aided Systems Theory - EUROCAST 2022: 18th International Conference, Las Palmas de Gran Canaria, Feb. 2022}.
}
\renewcommand\thefootnote{\arabic{footnote}}

\begin{abstract}

Fast Function Extraction (FFX) is a deterministic algorithm for solving symbolic regression problems. We improve the accuracy of FFX by adding parameters to the arguments of nonlinear functions. Instead of only optimizing linear parameters, we optimize these additional nonlinear parameters with separable nonlinear least squared optimization using a variable projection algorithm. Both FFX and our new algorithm is applied on the PennML benchmark suite. We show that the proposed extensions of FFX leads to higher accuracy while providing models of similar length and with only a small increase in runtime on the given data. Our results are compared to a large set of regression methods that were already published for the given benchmark suite.

\end{abstract}

\section{Symbolic Regression and FFX}

Symbolic regression is a machine learning task in which we try to identify mathematical formulas that cover linear and nonlinear relations within given data. The most common algorithm for solving symbolic regression is genetic programming (GP), which optimizes a population of mathematical models using crossover and mutation. GP is in theory capable of finding models of any syntactical structure and complexity. However, disadvantages of GP are its stochasticity, its long algorithm runtime for nontrivial problems and its complex hyperparameter settings. These characteristics led to the development of non-evolutionary algorithms which produce more restricted models but have advantages in determinism, runtime or complexity of hyperparameters \cite{de2018greedy,kammerer2020symbolic,mcconaghy2011ffx}.

One of the first algorithms that was developed to tackle the shortages of GP is Fast Function Extraction (FFX) \cite{mcconaghy2011ffx}. FFX generates a large set of \textit{base functions} first, then it learns a regularized linear model using these base functions as terms. The set of base functions is a combination of the original features with several nonlinear functions such as $\exp(\dots)$ or $\log(\dots)$ and a predefined set of real-valued exponent values. Examples of base functions for problems with features $\{x_1, x_2\}$ are $x_1$, $x_2$, $x_1{}^2$, $\exp(x_1)$, $\exp(x_1{}^2)$, $\log(x_1)$ or $\log(x_2{}^{0.5})$.

The structure of FFX models with $n$ features $\{x_1...x_n\}$ is outlined in Equation \ref{eq:ffx_sample}. Parameters $\{c_0, c_1, $ $\dots, c_m\}$ are linear parameters of a model with $m$ base functions. They are learned with ElasticNet regression \cite{friedman2017elements}. ElasticNet regression identifies only the most relevant base functions due to its regularization by setting linear parameters $c_i$ of irrelevant base functions to zero (sparsification). Therefore, learned models contain only a subset of the original base functions.

\begin{equation}
  \begin{aligned}
  \label{eq:ffx_sample}
  \hat{f}(\mathbf{x}) = c_0 + c_1 & \ \text{func}_1(x_1{}^{e_1}) + \dots + c_m \ \text{func}_m(x_n{}^{e_m}) \\
  \text{with} \ \text{func}_1, \dots, \text{func}_m \in & \{\text{abs}(), \log(), \dots\} \ \text{and} \ e_1, \dots, e_m \in \{0.5, 1, 2 \}
  \end{aligned}
\end{equation}

\subsection{Motivation and Objectives}

A disadvantage of FFX is that only linear parameters are optimized. Parameters within nonlinear functions are not present. For example functions with feature $x$ and nonlinear parameters $k_i$ such as $\log(x + k_i)$ or $\exp(k_i x)$ have to be approximated by FFX with a linear model of several base functions. In this work, we extend the capabilities of FFX by adding such a real-valued parameter $k$ to each generated base functions. Adding several nonlinear parameters to the possible model structure should allow to fit additive models with fewer base functions, as we have more degrees of freedom per base function.

The introduced nonlinear parameters are optimized in combination with the linear parameters by separable nonlinear least squares optmization (NLS) using a variable projection algorithm~\cite{chen2018regularized}. We call this new algorithm \textit{FFX with Nonlinear Least Squares Optimization} (FFX NLS). We test whether FFX NLS leads to higher accuracy than the original implementation on the \textit{PennML} benchmark suite \cite{orzechowski2018we} and compare the complexity of the generated models. Given, that we have more degrees of freedom to fit data with a single base function, we expect to find that it produces models with a lower number of base functions and therefore simpler models than the original FFX algorithm.

\section{Algorithm Description}

Similar to FFX, FFX NLS runs in several steps. First, base functions are generated. Then the most relevant base functions and the model's parameters are determined. In difference to FFX, the selection of relevant base function and the optimization of parameters are separate steps. FFX NLS performs the following four steps that are described in detail in the next sections:

\begin{enumerate}
    \item Generate a list of all univariate base functions $\boldsymbol{f}$ (cf.~Section \ref{subsec:baseFunctions}).
    \item Optimize all parameters $\boldsymbol{k}$ and $\boldsymbol{l}$ of a nonlinear model $l_0 + \sum_{i}{l_i f_i(\boldsymbol{k}, \boldsymbol{x})}$ that consists of the base functions as terms. $\boldsymbol{k}$ is a vector of all nonlinear parameters. $\boldsymbol{l}$ are linear parameters, $\boldsymbol{x}$ are features (cf.~Section \ref{subsec:ParameterOptimization}). 
    \item Select most important base functions of $\boldsymbol{f}$ with a regularized linear model and the nonlinear parameters $\boldsymbol{k}$ from the previous step (cf.~Section \ref{subsec:BaseFunctionSelection}).
    \item The final model is created by optimizing all parameters again with nonlinear least squares optimization but only using the most important base function (cf.~Section \ref{subsec:finalModel}).
\end{enumerate}

\subsection{Base Functions}
\label{subsec:baseFunctions}

In step 1, we generate all univariate base functions $\boldsymbol{f}$ with placeholders for nonlinear parameters. For each feature $x_i$ we create base functions of structure $\mathit{func}(a x_i^p + b)$ with $\mathit{func} \in \{\text{id}, \log, \exp, \text{sqrt}\}$, $p \in \{1, 2\}$ and two scaling values $a$ and $b$ with $\text{id}(x) = x$. The scaling values $a$ and $b$ are placeholder for nonlinear parameters that will be optimized later on. We also include bivariate base functions, which are described in Section \ref{subsec:finalModel}.

We utilize the linear structure in the final model as well as mathematical identities of the used nonlinear functions to reduce the number of nonlinear parameters. Due to mathematical identities such as $\log(ax_i) = \log(a) + \log(x_i)$ with $a$ being a parameter that is trained later on, we can skip certain scaling values. In the case of logarithm, we can just use $\log(x_i + b)$ instead of $\log(a x_i + b)$ as base function as we can rewrite it to $log(c) \log(x_i + d)$. Then we can skip $log(a)$ as it is constant in the final model and therefore summed up by the final model's intercept. The same applies to the exp-function, in which we can skip the multiplicative scaling value in the argument as we can rewrite $\exp(x+a)=\exp(x)\exp(a)$ and skip $\exp(a)$ in the final model.

\subsection{Parameter Optimization with Variable Projection}
\label{subsec:ParameterOptimization}

The generated base functions $\boldsymbol{f}$ are combined to one large linear model $\hat{\Theta}(\boldsymbol{x}) = l_0 + \sum_{i}{l_i f_i(\boldsymbol{k}, \boldsymbol{x})}$ with $\boldsymbol{l}$ as vector of linear parameters and intercept and $\boldsymbol{k}$ as vector of nonlinear parameters. We use NLS to find the values in $\boldsymbol{l}$ and $\boldsymbol{k}$ that minimize the mean squared error (MSE) for given training data.

Since nonlinear least squares optimization is computationally expensive, we use a variable projection (VP) algorithm initially developed by Golub and Pereya \cite{golub1973differentiation} for optimizing all parameters $\boldsymbol{l}$ and $\boldsymbol{k}$. The advantage of VP over plain NLS optimization is that VP utilizes the generated model's structure -- a nonlinear model with several linear parameters $\boldsymbol{l}$. Golub and Pereya call the given model structure a \textit{separable least squares problem}. This setting is common in engineering domains. VP optimizes nonlinear parameters in $\boldsymbol{k}$ iteratively, while optimal linear parameters are solved exactly via ordinary least squares. Given the high number of linear parameters in the model (due to the large number of base functions), the use of VP is an important performance aspect of FFX NLS. We use the effieicennt VP algorithm by Krogh \cite{krogh1974efficient}.

\subsection{Base Function Selection}
\label{subsec:BaseFunctionSelection}

As we generate a very large number of base functions, we need to select the most relevant ones in order to create a both interpretable and well-generalizing model. The number of selected base functions is thereby a hyperparameter of FFX NLS. In plain FFX, the selection of relevant base functions and the optimization of parameters is done in one step with a ElasticNet regression \cite{friedman2017elements}, as only linear parameters need to be optimized.

Since we also need to optimize nonlinear parameters, no simple way to combine NLS optimization and regularization of linear parameters is available. The VP algorithm by Chen et al.~\cite{chen2018regularized} uses Tikhonov regularization \cite{friedman2017elements}. However, in initial experiments this algorithm was not beneficial for FFX NLS to identify relevant terms. Tikhonov regularization shrinks linear parameters, however, it did not provide the necessary sparsification of linear parameters and the algorithm in \cite{chen2018regularized} is computationally more expensive. Alternatively, we use the already computed vector of nonlinear parameters $\boldsymbol{k}$ of the previous step as fixed constants and optimize the linear parameters $\boldsymbol{l}$ with a lasso regression \cite{friedman2017elements}. Terms with a linear parameter $\neq 0$ are selected as most important base functions. We get the desired number of base functions by iteratively increasing the lasso regression's $\lambda$ value in order to shrink more parameters to zero. Although this method ignores dependencies between linear and nonlinear parameters, it is still effective for selecting base functions both regarding runtime and further modelling accuracy.

\subsection{Training of Final Model and Bivariate Base Functions}
\label{subsec:finalModel}

Similar to the parameter optimization in Section \ref{subsec:ParameterOptimization}, we combine base functions to one single model and optimize all parameters again. In difference to the previous NLS optimization step, we take in this step only the most relevant base functions from Section \ref{subsec:BaseFunctionSelection}. As parameters are not independent from each other, we repeat this step to get optimal parameters for a subset of base functions.

To cover interactions between features, we also include bivariate base functions. For this we take pairwise combinations of the previously generated univariate base function and multiply them to get one new base function for each pair. To prevent a combinatorial explosion of base functions and nonlinear parameters in highly multidimensional problems, we only consider the ten most important univariate base functions from Section \ref{subsec:BaseFunctionSelection} for to create bivariate base functions. We take the resulting ${10 \choose 2 }= 45$ bivariate functions and append them to our existing list of univariate base functions. Then we repeat the steps described in Section \ref{subsec:ParameterOptimization} and \ref{subsec:BaseFunctionSelection} to determine the most important base functions across both univariate and bivariate base function for the final model. We consider only ten univariate base functions as a larger number did not provide any benefit in achieving our objectives.

\section{Experimental Setup}

We use \textit{PennML} benchmark suite \cite{orzechowski2018we} for all experiments. This benchmark consist of over 90 regression problem and provides a performance overview of several common regression algorithms. We apply both FFX and FFX NLS on these problems to compare the algorithms' accuracies with each other and with other common regression algorithms. Thereby we take the experimental results from \cite{kommenda2020parameter} for a comprehensive comparison as this work shows the results of the currently most accurate regression algorithms like GP CoOp as implemented in \cite{burlacu2020operon} for the PennML data.

We apply the same modelling workflow as in \cite{orzechowski2018we} in our experiments. For every regression problem in the benchmark suite, we repeat the modelling ten times. We shuffle the dataset in every repetition and take the first 75\% of observations as training set and the remaining ones as test set. We perform a grid search with a 5-fold cross validation for each shuffled dataset for hyperparameter tuning and train one final model with these best hyperparameters. Eventually, we get ten different models for each original regression problem. Hyperparameters for all other algorithms are described in \cite{kommenda2020parameter}. We use the following hyperparameter sets for both FFX and FFX NLS:

\begin{itemize}
    \item Max. number of base function in final model $\in \{3, 5, 10, 20, 30, 50\}$
    \item Use bivariate base functions $\in \{\text{true}, \text{false}\}$
    \item Use nonlinear functions $\in \{\text{true}, \text{false}\}$
    \item FFX only: L1 ratio $\in \{0, 0.5, 1\}$
\end{itemize}

\section{Results}

\begin{figure}[b!]
    \begin{center}
        \input{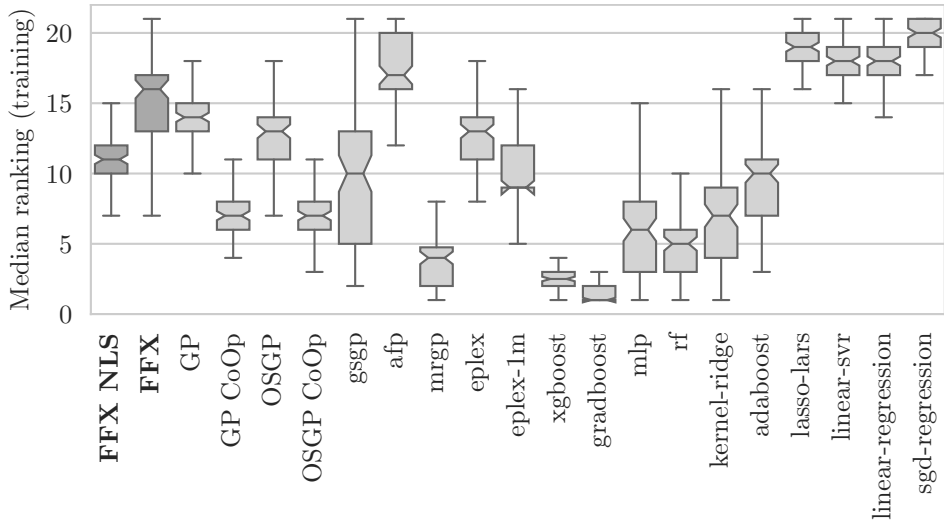}
    \end{center}
    \caption{Distribution of median rankings of the mean squared error on the training set.}
    \label{fig:training}
\end{figure}

\begin{figure}[t!]
    \begin{center}
        \input{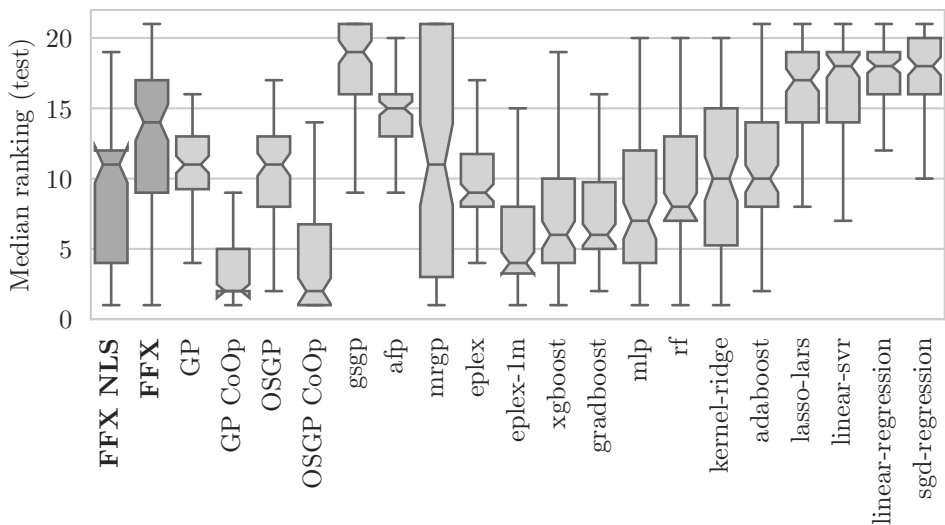}
    \end{center}
    \caption{Distribution of median rankings of the mean squared error on the test set.}
    \label{fig:test}
\end{figure}

As in the original experiments for this benchmark suite \cite{orzechowski2018we}, we first calculate the median (mean squared) error of the ten models of each regression problem. Then we rank all regression methods by their median error for each problem. Figure \ref{fig:training} and \ref{fig:test} show the distribution of ranks for each algorithm for training and test across all regression problems in the benchmark suite. E.g.~ in Figure \ref{fig:training}, gradient boosting was for most problems the most accurate algorithm in training.

Figure \ref{fig:training} and \ref{fig:test} show the distribution of rankings of each algorithm's median rank on the training set and test set. The optimization of nonlinear parameters in FFX NLS provided to more accurate models than FFX both in training and test. In comparison to other algorithms, FFX as well as FFX NLS perform better than linear models, which are on the right of Figure \ref{fig:training} (like linear regression or lasso regression). This is expected given both FFX methods can cover nonlinear dependencies and interactions in the data in contrast to purely linear methods. However, both algorithms perform worse than boosting methods or GP with NLS (GP CoOp) by Kommenda et al.~\cite{kommenda2020parameter}. Also this is plausible, as both methods have a more powerful search algorithm and a larger hypothesis space than FFX methods with their restricted model structures.

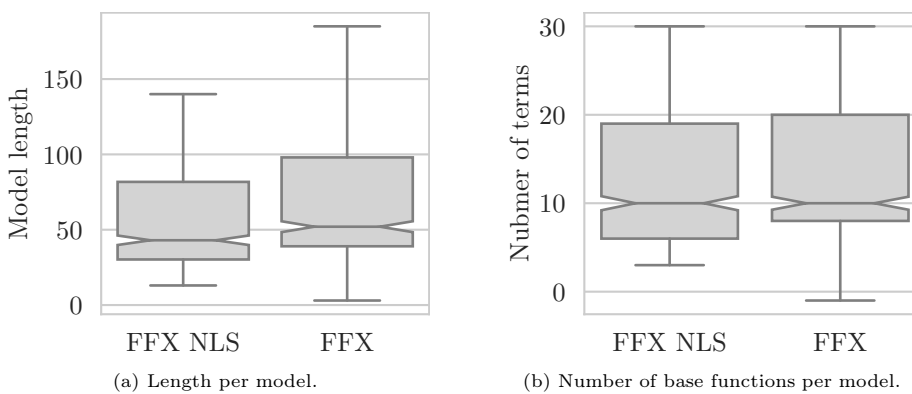
\begin{figure}[t!]
    \centering
    \begin{subfigure}[b]{0.45\textwidth}
        \centering
        \input{plots/tree_length.pgf}
        \caption{Length per model.}
        \label{fig:length}
    \end{subfigure}
    \hfill
    \begin{subfigure}[b]{0.45\textwidth}
        \centering
        \input{plots/terms.pgf}
        \caption{Number of base functions per model.}
        \label{fig:terms}
    \end{subfigure}
    \vspace{12pt}
    \caption{Distribution complexity of all models from FFX and FFX NLS.}
    \label{fig:complexity}
\end{figure}

To analyze the size of models, we count the number of syntactical symbols in a model. E.g.~the model $c_0 + \log{x_1 + c_2}$ has a complexity of six. We use this measure because it takes the added scaling terms within function arguments in FFX NLS into account. Figure \ref{fig:length} shows that both the size of models and the number of base function within models produced by FFX NLS and FFX are similar. FFX NLS models are just slightly shorter.

\begin{figure}[b!]
    \begin{center}
        \input{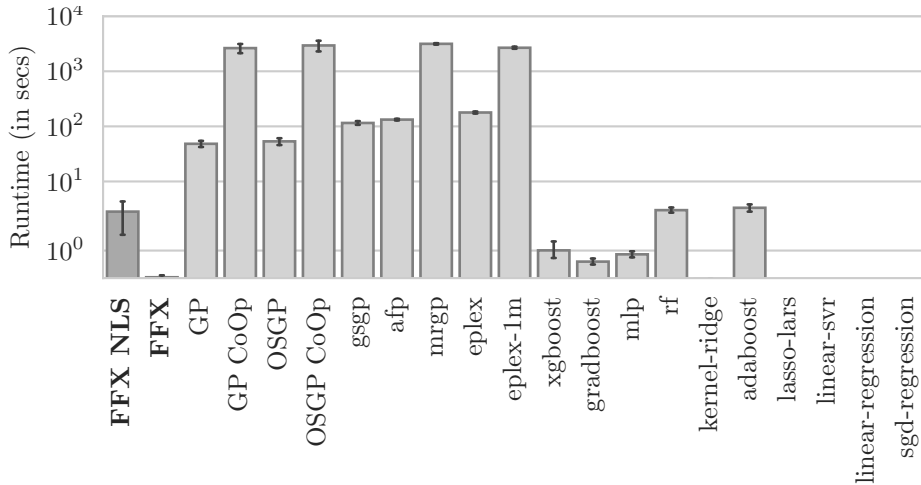}
    \end{center}
    \caption{Median runtime of all algorithms.}
    \label{fig:runtime}
\end{figure} 

Figure \ref{fig:runtime} shows the median runtime across all problems per algorithm. While one FFX run takes less than a second for most problems, the runtime of FFX NLS ranges between one and a few seconds. Although the runtime increase is relatively large, it is still feasible to perform large grid searches in reasonable amount of time with FFX NLS. Both algorithms beat some GP-based algorithms and are on a similar level as boosting algorithms.

\section{Conclusion}

We proposed an extension of the model structure of the FFX algorithm. We added nonlinear scaling parameter which were optimized using the variable projection algorithm by Krogh \cite{krogh1974efficient} and separate base function selection, which we called FFX NLS. We achieved a large improvement in test accuracy in comparison to plain FFX on the PennML benchmark suite while providing models of similar complexity. Although the runtime of FFX NLS is higher than the one of FFX, it still finishes training within seconds.

However, FFX and FFX NLS still perform worse than many symbolic regression algorithms. Big advantages of FFX NLS towards GP-based algorithms are its short runtime, the low number of hyperparameters and the simple, comprehensible model structure. Compared to boosting algorithms, FFX NLS performs worse in accuracy but similar in runtime. However, boosting algorithms are considered black box methods and allow no readability of its models. 

To sum it up, FFX is a promising tool for quick data exploration. It identifies interactions as well as nonlinear relations with simple hyperparameter configuration and quick execution. However, further improvements regarding accuracy are needed. Potential improvements are the combination of regularization and parameter optimization, as this are separate steps right now that deliberately ignore dependencies between linear and nonlinear parameters.

\subsubsection{Acknowledgements} The authors gratefully acknowledge support by the Christian Doppler Research Association and the Federal Ministry for Digital and Economic Affairs within the \emph{Josef Ressel Center for Symbolic Regression}.

%
%
%
\bibliographystyle{splncs04}
\bibliography{ref}

\end{document}

%% file: plots/tree_length.pgf
\begingroup%
\makeatletter%
\begin{pgfpicture}%
\pgfpathrectangle{\pgfpointorigin}{\pgfqpoint{2.201240in}{1.800461in}}%
\pgfusepath{use as bounding box, clip}%
\begin{pgfscope}%
\pgfsetbuttcap%
\pgfsetmiterjoin%
\pgfsetlinewidth{0.000000pt}%
\definecolor{currentstroke}{rgb}{1.000000,1.000000,1.000000}%
\pgfsetstrokecolor{currentstroke}%
\pgfsetstrokeopacity{0.000000}%
\pgfsetdash{}{0pt}%
\pgfpathmoveto{\pgfqpoint{0.000000in}{0.000000in}}%
\pgfpathlineto{\pgfqpoint{2.201240in}{0.000000in}}%
\pgfpathlineto{\pgfqpoint{2.201240in}{1.800461in}}%
\pgfpathlineto{\pgfqpoint{0.000000in}{1.800461in}}%
\pgfpathlineto{\pgfqpoint{0.000000in}{0.000000in}}%
\pgfpathclose%
\pgfusepath{}%
\end{pgfscope}%
\begin{pgfscope}%
\pgfsetbuttcap%
\pgfsetmiterjoin%
\definecolor{currentfill}{rgb}{1.000000,1.000000,1.000000}%
\pgfsetfillcolor{currentfill}%
\pgfsetlinewidth{0.000000pt}%
\definecolor{currentstroke}{rgb}{0.000000,0.000000,0.000000}%
\pgfsetstrokecolor{currentstroke}%
\pgfsetstrokeopacity{0.000000}%
\pgfsetdash{}{0pt}%
\pgfpathmoveto{\pgfqpoint{0.484568in}{0.220679in}}%
\pgfpathlineto{\pgfqpoint{2.201240in}{0.220679in}}%
\pgfpathlineto{\pgfqpoint{2.201240in}{1.800461in}}%
\pgfpathlineto{\pgfqpoint{0.484568in}{1.800461in}}%
\pgfpathlineto{\pgfqpoint{0.484568in}{0.220679in}}%
\pgfpathclose%
\pgfusepath{fill}%
\end{pgfscope}%
\begin{pgfscope}%
\definecolor{textcolor}{rgb}{0.150000,0.150000,0.150000}%
\pgfsetstrokecolor{textcolor}%
\pgfsetfillcolor{textcolor}%
\pgftext[x=0.913736in,y=0.123457in,,top]{\color{textcolor}\rmfamily\fontsize{10.000000}{12.000000}\selectfont FFX NLS}%
\end{pgfscope}%
\begin{pgfscope}%
\definecolor{textcolor}{rgb}{0.150000,0.150000,0.150000}%
\pgfsetstrokecolor{textcolor}%
\pgfsetfillcolor{textcolor}%
\pgftext[x=1.772072in,y=0.123457in,,top]{\color{textcolor}\rmfamily\fontsize{10.000000}{12.000000}\selectfont FFX}%
\end{pgfscope}%
\begin{pgfscope}%
\pgfpathrectangle{\pgfqpoint{0.484568in}{0.220679in}}{\pgfqpoint{1.716671in}{1.579782in}}%
\pgfusepath{clip}%
\pgfsetroundcap%
\pgfsetroundjoin%
\pgfsetlinewidth{0.803000pt}%
\definecolor{currentstroke}{rgb}{0.800000,0.800000,0.800000}%
\pgfsetstrokecolor{currentstroke}%
\pgfsetdash{}{0pt}%
\pgfpathmoveto{\pgfqpoint{0.484568in}{0.268814in}}%
\pgfpathlineto{\pgfqpoint{2.201240in}{0.268814in}}%
\pgfusepath{stroke}%
\end{pgfscope}%
\begin{pgfscope}%
\definecolor{textcolor}{rgb}{0.150000,0.150000,0.150000}%
\pgfsetstrokecolor{textcolor}%
\pgfsetfillcolor{textcolor}%
\pgftext[x=0.317902in, y=0.220589in, left, base]{\color{textcolor}\rmfamily\fontsize{10.000000}{12.000000}\selectfont \(\displaystyle {0}\)}%
\end{pgfscope}%
\begin{pgfscope}%
\pgfpathrectangle{\pgfqpoint{0.484568in}{0.220679in}}{\pgfqpoint{1.716671in}{1.579782in}}%
\pgfusepath{clip}%
\pgfsetroundcap%
\pgfsetroundjoin%
\pgfsetlinewidth{0.803000pt}%
\definecolor{currentstroke}{rgb}{0.800000,0.800000,0.800000}%
\pgfsetstrokecolor{currentstroke}%
\pgfsetdash{}{0pt}%
\pgfpathmoveto{\pgfqpoint{0.484568in}{0.663365in}}%
\pgfpathlineto{\pgfqpoint{2.201240in}{0.663365in}}%
\pgfusepath{stroke}%
\end{pgfscope}%
\begin{pgfscope}%
\definecolor{textcolor}{rgb}{0.150000,0.150000,0.150000}%
\pgfsetstrokecolor{textcolor}%
\pgfsetfillcolor{textcolor}%
\pgftext[x=0.248457in, y=0.615140in, left, base]{\color{textcolor}\rmfamily\fontsize{10.000000}{12.000000}\selectfont \(\displaystyle {50}\)}%
\end{pgfscope}%
\begin{pgfscope}%
\pgfpathrectangle{\pgfqpoint{0.484568in}{0.220679in}}{\pgfqpoint{1.716671in}{1.579782in}}%
\pgfusepath{clip}%
\pgfsetroundcap%
\pgfsetroundjoin%
\pgfsetlinewidth{0.803000pt}%
\definecolor{currentstroke}{rgb}{0.800000,0.800000,0.800000}%
\pgfsetstrokecolor{currentstroke}%
\pgfsetdash{}{0pt}%
\pgfpathmoveto{\pgfqpoint{0.484568in}{1.057916in}}%
\pgfpathlineto{\pgfqpoint{2.201240in}{1.057916in}}%
\pgfusepath{stroke}%
\end{pgfscope}%
\begin{pgfscope}%
\definecolor{textcolor}{rgb}{0.150000,0.150000,0.150000}%
\pgfsetstrokecolor{textcolor}%
\pgfsetfillcolor{textcolor}%
\pgftext[x=0.179012in, y=1.009691in, left, base]{\color{textcolor}\rmfamily\fontsize{10.000000}{12.000000}\selectfont \(\displaystyle {100}\)}%
\end{pgfscope}%
\begin{pgfscope}%
\pgfpathrectangle{\pgfqpoint{0.484568in}{0.220679in}}{\pgfqpoint{1.716671in}{1.579782in}}%
\pgfusepath{clip}%
\pgfsetroundcap%
\pgfsetroundjoin%
\pgfsetlinewidth{0.803000pt}%
\definecolor{currentstroke}{rgb}{0.800000,0.800000,0.800000}%
\pgfsetstrokecolor{currentstroke}%
\pgfsetdash{}{0pt}%
\pgfpathmoveto{\pgfqpoint{0.484568in}{1.452467in}}%
\pgfpathlineto{\pgfqpoint{2.201240in}{1.452467in}}%
\pgfusepath{stroke}%
\end{pgfscope}%
\begin{pgfscope}%
\definecolor{textcolor}{rgb}{0.150000,0.150000,0.150000}%
\pgfsetstrokecolor{textcolor}%
\pgfsetfillcolor{textcolor}%
\pgftext[x=0.179012in, y=1.404242in, left, base]{\color{textcolor}\rmfamily\fontsize{10.000000}{12.000000}\selectfont \(\displaystyle {150}\)}%
\end{pgfscope}%
\begin{pgfscope}%
\definecolor{textcolor}{rgb}{0.150000,0.150000,0.150000}%
\pgfsetstrokecolor{textcolor}%
\pgfsetfillcolor{textcolor}%
\pgftext[x=0.123457in,y=1.010570in,,bottom,rotate=90.000000]{\color{textcolor}\rmfamily\fontsize{10.000000}{12.000000}\selectfont Model length}%
\end{pgfscope}%
\begin{pgfscope}%
\pgfpathrectangle{\pgfqpoint{0.484568in}{0.220679in}}{\pgfqpoint{1.716671in}{1.579782in}}%
\pgfusepath{clip}%
\pgfsetbuttcap%
\pgfsetmiterjoin%
\definecolor{currentfill}{rgb}{0.827451,0.827451,0.827451}%
\pgfsetfillcolor{currentfill}%
\pgfsetlinewidth{1.003750pt}%
\definecolor{currentstroke}{rgb}{0.498039,0.498039,0.498039}%
\pgfsetstrokecolor{currentstroke}%
\pgfsetdash{}{0pt}%
\pgfpathmoveto{\pgfqpoint{0.570402in}{0.507517in}}%
\pgfpathlineto{\pgfqpoint{1.257070in}{0.507517in}}%
\pgfpathlineto{\pgfqpoint{1.257070in}{0.583255in}}%
\pgfpathlineto{\pgfqpoint{1.085403in}{0.608128in}}%
\pgfpathlineto{\pgfqpoint{1.257070in}{0.633001in}}%
\pgfpathlineto{\pgfqpoint{1.257070in}{0.913905in}}%
\pgfpathlineto{\pgfqpoint{0.570402in}{0.913905in}}%
\pgfpathlineto{\pgfqpoint{0.570402in}{0.633001in}}%
\pgfpathlineto{\pgfqpoint{0.742069in}{0.608128in}}%
\pgfpathlineto{\pgfqpoint{0.570402in}{0.583255in}}%
\pgfpathlineto{\pgfqpoint{0.570402in}{0.507517in}}%
\pgfpathlineto{\pgfqpoint{0.570402in}{0.507517in}}%
\pgfpathclose%
\pgfusepath{stroke,fill}%
\end{pgfscope}%
\begin{pgfscope}%
\pgfpathrectangle{\pgfqpoint{0.484568in}{0.220679in}}{\pgfqpoint{1.716671in}{1.579782in}}%
\pgfusepath{clip}%
\pgfsetbuttcap%
\pgfsetmiterjoin%
\definecolor{currentfill}{rgb}{0.827451,0.827451,0.827451}%
\pgfsetfillcolor{currentfill}%
\pgfsetlinewidth{1.003750pt}%
\definecolor{currentstroke}{rgb}{0.498039,0.498039,0.498039}%
\pgfsetstrokecolor{currentstroke}%
\pgfsetdash{}{0pt}%
\pgfpathmoveto{\pgfqpoint{1.428738in}{0.576564in}}%
\pgfpathlineto{\pgfqpoint{2.115406in}{0.576564in}}%
\pgfpathlineto{\pgfqpoint{2.115406in}{0.650630in}}%
\pgfpathlineto{\pgfqpoint{1.943739in}{0.679147in}}%
\pgfpathlineto{\pgfqpoint{2.115406in}{0.707664in}}%
\pgfpathlineto{\pgfqpoint{2.115406in}{1.042134in}}%
\pgfpathlineto{\pgfqpoint{1.428738in}{1.042134in}}%
\pgfpathlineto{\pgfqpoint{1.428738in}{0.707664in}}%
\pgfpathlineto{\pgfqpoint{1.600405in}{0.679147in}}%
\pgfpathlineto{\pgfqpoint{1.428738in}{0.650630in}}%
\pgfpathlineto{\pgfqpoint{1.428738in}{0.576564in}}%
\pgfpathlineto{\pgfqpoint{1.428738in}{0.576564in}}%
\pgfpathclose%
\pgfusepath{stroke,fill}%
\end{pgfscope}%
\begin{pgfscope}%
\pgfpathrectangle{\pgfqpoint{0.484568in}{0.220679in}}{\pgfqpoint{1.716671in}{1.579782in}}%
\pgfusepath{clip}%
\pgfsetroundcap%
\pgfsetroundjoin%
\pgfsetlinewidth{1.003750pt}%
\definecolor{currentstroke}{rgb}{0.498039,0.498039,0.498039}%
\pgfsetstrokecolor{currentstroke}%
\pgfsetdash{}{0pt}%
\pgfpathmoveto{\pgfqpoint{0.913736in}{0.507517in}}%
\pgfpathlineto{\pgfqpoint{0.913736in}{0.371397in}}%
\pgfusepath{stroke}%
\end{pgfscope}%
\begin{pgfscope}%
\pgfpathrectangle{\pgfqpoint{0.484568in}{0.220679in}}{\pgfqpoint{1.716671in}{1.579782in}}%
\pgfusepath{clip}%
\pgfsetroundcap%
\pgfsetroundjoin%
\pgfsetlinewidth{1.003750pt}%
\definecolor{currentstroke}{rgb}{0.498039,0.498039,0.498039}%
\pgfsetstrokecolor{currentstroke}%
\pgfsetdash{}{0pt}%
\pgfpathmoveto{\pgfqpoint{0.913736in}{0.913905in}}%
\pgfpathlineto{\pgfqpoint{0.913736in}{1.373557in}}%
\pgfusepath{stroke}%
\end{pgfscope}%
\begin{pgfscope}%
\pgfpathrectangle{\pgfqpoint{0.484568in}{0.220679in}}{\pgfqpoint{1.716671in}{1.579782in}}%
\pgfusepath{clip}%
\pgfsetroundcap%
\pgfsetroundjoin%
\pgfsetlinewidth{1.003750pt}%
\definecolor{currentstroke}{rgb}{0.498039,0.498039,0.498039}%
\pgfsetstrokecolor{currentstroke}%
\pgfsetdash{}{0pt}%
\pgfpathmoveto{\pgfqpoint{0.742069in}{0.371397in}}%
\pgfpathlineto{\pgfqpoint{1.085403in}{0.371397in}}%
\pgfusepath{stroke}%
\end{pgfscope}%
\begin{pgfscope}%
\pgfpathrectangle{\pgfqpoint{0.484568in}{0.220679in}}{\pgfqpoint{1.716671in}{1.579782in}}%
\pgfusepath{clip}%
\pgfsetroundcap%
\pgfsetroundjoin%
\pgfsetlinewidth{1.003750pt}%
\definecolor{currentstroke}{rgb}{0.498039,0.498039,0.498039}%
\pgfsetstrokecolor{currentstroke}%
\pgfsetdash{}{0pt}%
\pgfpathmoveto{\pgfqpoint{0.742069in}{1.373557in}}%
\pgfpathlineto{\pgfqpoint{1.085403in}{1.373557in}}%
\pgfusepath{stroke}%
\end{pgfscope}%
\begin{pgfscope}%
\pgfpathrectangle{\pgfqpoint{0.484568in}{0.220679in}}{\pgfqpoint{1.716671in}{1.579782in}}%
\pgfusepath{clip}%
\pgfsetroundcap%
\pgfsetroundjoin%
\pgfsetlinewidth{1.003750pt}%
\definecolor{currentstroke}{rgb}{0.498039,0.498039,0.498039}%
\pgfsetstrokecolor{currentstroke}%
\pgfsetdash{}{0pt}%
\pgfpathmoveto{\pgfqpoint{1.772072in}{0.576564in}}%
\pgfpathlineto{\pgfqpoint{1.772072in}{0.292487in}}%
\pgfusepath{stroke}%
\end{pgfscope}%
\begin{pgfscope}%
\pgfpathrectangle{\pgfqpoint{0.484568in}{0.220679in}}{\pgfqpoint{1.716671in}{1.579782in}}%
\pgfusepath{clip}%
\pgfsetroundcap%
\pgfsetroundjoin%
\pgfsetlinewidth{1.003750pt}%
\definecolor{currentstroke}{rgb}{0.498039,0.498039,0.498039}%
\pgfsetstrokecolor{currentstroke}%
\pgfsetdash{}{0pt}%
\pgfpathmoveto{\pgfqpoint{1.772072in}{1.042134in}}%
\pgfpathlineto{\pgfqpoint{1.772072in}{1.728653in}}%
\pgfusepath{stroke}%
\end{pgfscope}%
\begin{pgfscope}%
\pgfpathrectangle{\pgfqpoint{0.484568in}{0.220679in}}{\pgfqpoint{1.716671in}{1.579782in}}%
\pgfusepath{clip}%
\pgfsetroundcap%
\pgfsetroundjoin%
\pgfsetlinewidth{1.003750pt}%
\definecolor{currentstroke}{rgb}{0.498039,0.498039,0.498039}%
\pgfsetstrokecolor{currentstroke}%
\pgfsetdash{}{0pt}%
\pgfpathmoveto{\pgfqpoint{1.600405in}{0.292487in}}%
\pgfpathlineto{\pgfqpoint{1.943739in}{0.292487in}}%
\pgfusepath{stroke}%
\end{pgfscope}%
\begin{pgfscope}%
\pgfpathrectangle{\pgfqpoint{0.484568in}{0.220679in}}{\pgfqpoint{1.716671in}{1.579782in}}%
\pgfusepath{clip}%
\pgfsetroundcap%
\pgfsetroundjoin%
\pgfsetlinewidth{1.003750pt}%
\definecolor{currentstroke}{rgb}{0.498039,0.498039,0.498039}%
\pgfsetstrokecolor{currentstroke}%
\pgfsetdash{}{0pt}%
\pgfpathmoveto{\pgfqpoint{1.600405in}{1.728653in}}%
\pgfpathlineto{\pgfqpoint{1.943739in}{1.728653in}}%
\pgfusepath{stroke}%
\end{pgfscope}%
\begin{pgfscope}%
\pgfpathrectangle{\pgfqpoint{0.484568in}{0.220679in}}{\pgfqpoint{1.716671in}{1.579782in}}%
\pgfusepath{clip}%
\pgfsetroundcap%
\pgfsetroundjoin%
\pgfsetlinewidth{1.003750pt}%
\definecolor{currentstroke}{rgb}{0.498039,0.498039,0.498039}%
\pgfsetstrokecolor{currentstroke}%
\pgfsetdash{}{0pt}%
\pgfpathmoveto{\pgfqpoint{0.742069in}{0.608128in}}%
\pgfpathlineto{\pgfqpoint{1.085403in}{0.608128in}}%
\pgfusepath{stroke}%
\end{pgfscope}%
\begin{pgfscope}%
\pgfpathrectangle{\pgfqpoint{0.484568in}{0.220679in}}{\pgfqpoint{1.716671in}{1.579782in}}%
\pgfusepath{clip}%
\pgfsetroundcap%
\pgfsetroundjoin%
\pgfsetlinewidth{1.003750pt}%
\definecolor{currentstroke}{rgb}{0.498039,0.498039,0.498039}%
\pgfsetstrokecolor{currentstroke}%
\pgfsetdash{}{0pt}%
\pgfpathmoveto{\pgfqpoint{1.600405in}{0.679147in}}%
\pgfpathlineto{\pgfqpoint{1.943739in}{0.679147in}}%
\pgfusepath{stroke}%
\end{pgfscope}%
\begin{pgfscope}%
\pgfsetrectcap%
\pgfsetmiterjoin%
\pgfsetlinewidth{0.803000pt}%
\definecolor{currentstroke}{rgb}{0.800000,0.800000,0.800000}%
\pgfsetstrokecolor{currentstroke}%
\pgfsetdash{}{0pt}%
\pgfpathmoveto{\pgfqpoint{0.484568in}{0.220679in}}%
\pgfpathlineto{\pgfqpoint{0.484568in}{1.800461in}}%
\pgfusepath{stroke}%
\end{pgfscope}%
\begin{pgfscope}%
\pgfsetrectcap%
\pgfsetmiterjoin%
\pgfsetlinewidth{0.803000pt}%
\definecolor{currentstroke}{rgb}{0.800000,0.800000,0.800000}%
\pgfsetstrokecolor{currentstroke}%
\pgfsetdash{}{0pt}%
\pgfpathmoveto{\pgfqpoint{2.201240in}{0.220679in}}%
\pgfpathlineto{\pgfqpoint{2.201240in}{1.800461in}}%
\pgfusepath{stroke}%
\end{pgfscope}%
\begin{pgfscope}%
\pgfsetrectcap%
\pgfsetmiterjoin%
\pgfsetlinewidth{0.803000pt}%
\definecolor{currentstroke}{rgb}{0.800000,0.800000,0.800000}%
\pgfsetstrokecolor{currentstroke}%
\pgfsetdash{}{0pt}%
\pgfpathmoveto{\pgfqpoint{0.484568in}{0.220679in}}%
\pgfpathlineto{\pgfqpoint{2.201240in}{0.220679in}}%
\pgfusepath{stroke}%
\end{pgfscope}%
\begin{pgfscope}%
\pgfsetrectcap%
\pgfsetmiterjoin%
\pgfsetlinewidth{0.803000pt}%
\definecolor{currentstroke}{rgb}{0.800000,0.800000,0.800000}%
\pgfsetstrokecolor{currentstroke}%
\pgfsetdash{}{0pt}%
\pgfpathmoveto{\pgfqpoint{0.484568in}{1.800461in}}%
\pgfpathlineto{\pgfqpoint{2.201240in}{1.800461in}}%
\pgfusepath{stroke}%
\end{pgfscope}%
\end{pgfpicture}%
\makeatother%
\endgroup%

%% file: plots/terms.pgf
\begingroup%
\makeatletter%
\begin{pgfpicture}%
\pgfpathrectangle{\pgfpointorigin}{\pgfqpoint{2.200980in}{1.800461in}}%
\pgfusepath{use as bounding box, clip}%
\begin{pgfscope}%
\pgfsetbuttcap%
\pgfsetmiterjoin%
\pgfsetlinewidth{0.000000pt}%
\definecolor{currentstroke}{rgb}{1.000000,1.000000,1.000000}%
\pgfsetstrokecolor{currentstroke}%
\pgfsetstrokeopacity{0.000000}%
\pgfsetdash{}{0pt}%
\pgfpathmoveto{\pgfqpoint{0.000000in}{0.000000in}}%
\pgfpathlineto{\pgfqpoint{2.200980in}{0.000000in}}%
\pgfpathlineto{\pgfqpoint{2.200980in}{1.800461in}}%
\pgfpathlineto{\pgfqpoint{0.000000in}{1.800461in}}%
\pgfpathlineto{\pgfqpoint{0.000000in}{0.000000in}}%
\pgfpathclose%
\pgfusepath{}%
\end{pgfscope}%
\begin{pgfscope}%
\pgfsetbuttcap%
\pgfsetmiterjoin%
\definecolor{currentfill}{rgb}{1.000000,1.000000,1.000000}%
\pgfsetfillcolor{currentfill}%
\pgfsetlinewidth{0.000000pt}%
\definecolor{currentstroke}{rgb}{0.000000,0.000000,0.000000}%
\pgfsetstrokecolor{currentstroke}%
\pgfsetstrokeopacity{0.000000}%
\pgfsetdash{}{0pt}%
\pgfpathmoveto{\pgfqpoint{0.415124in}{0.220679in}}%
\pgfpathlineto{\pgfqpoint{2.200980in}{0.220679in}}%
\pgfpathlineto{\pgfqpoint{2.200980in}{1.800461in}}%
\pgfpathlineto{\pgfqpoint{0.415124in}{1.800461in}}%
\pgfpathlineto{\pgfqpoint{0.415124in}{0.220679in}}%
\pgfpathclose%
\pgfusepath{fill}%
\end{pgfscope}%
\begin{pgfscope}%
\definecolor{textcolor}{rgb}{0.150000,0.150000,0.150000}%
\pgfsetstrokecolor{textcolor}%
\pgfsetfillcolor{textcolor}%
\pgftext[x=0.861588in,y=0.123457in,,top]{\color{textcolor}\rmfamily\fontsize{10.000000}{12.000000}\selectfont FFX NLS}%
\end{pgfscope}%
\begin{pgfscope}%
\definecolor{textcolor}{rgb}{0.150000,0.150000,0.150000}%
\pgfsetstrokecolor{textcolor}%
\pgfsetfillcolor{textcolor}%
\pgftext[x=1.754516in,y=0.123457in,,top]{\color{textcolor}\rmfamily\fontsize{10.000000}{12.000000}\selectfont FFX}%
\end{pgfscope}%
\begin{pgfscope}%
\pgfpathrectangle{\pgfqpoint{0.415124in}{0.220679in}}{\pgfqpoint{1.785856in}{1.579782in}}%
\pgfusepath{clip}%
\pgfsetroundcap%
\pgfsetroundjoin%
\pgfsetlinewidth{0.803000pt}%
\definecolor{currentstroke}{rgb}{0.800000,0.800000,0.800000}%
\pgfsetstrokecolor{currentstroke}%
\pgfsetdash{}{0pt}%
\pgfpathmoveto{\pgfqpoint{0.415124in}{0.338815in}}%
\pgfpathlineto{\pgfqpoint{2.200980in}{0.338815in}}%
\pgfusepath{stroke}%
\end{pgfscope}%
\begin{pgfscope}%
\definecolor{textcolor}{rgb}{0.150000,0.150000,0.150000}%
\pgfsetstrokecolor{textcolor}%
\pgfsetfillcolor{textcolor}%
\pgftext[x=0.248457in, y=0.290590in, left, base]{\color{textcolor}\rmfamily\fontsize{10.000000}{12.000000}\selectfont \(\displaystyle {0}\)}%
\end{pgfscope}%
\begin{pgfscope}%
\pgfpathrectangle{\pgfqpoint{0.415124in}{0.220679in}}{\pgfqpoint{1.785856in}{1.579782in}}%
\pgfusepath{clip}%
\pgfsetroundcap%
\pgfsetroundjoin%
\pgfsetlinewidth{0.803000pt}%
\definecolor{currentstroke}{rgb}{0.800000,0.800000,0.800000}%
\pgfsetstrokecolor{currentstroke}%
\pgfsetdash{}{0pt}%
\pgfpathmoveto{\pgfqpoint{0.415124in}{0.802094in}}%
\pgfpathlineto{\pgfqpoint{2.200980in}{0.802094in}}%
\pgfusepath{stroke}%
\end{pgfscope}%
\begin{pgfscope}%
\definecolor{textcolor}{rgb}{0.150000,0.150000,0.150000}%
\pgfsetstrokecolor{textcolor}%
\pgfsetfillcolor{textcolor}%
\pgftext[x=0.179012in, y=0.753869in, left, base]{\color{textcolor}\rmfamily\fontsize{10.000000}{12.000000}\selectfont \(\displaystyle {10}\)}%
\end{pgfscope}%
\begin{pgfscope}%
\pgfpathrectangle{\pgfqpoint{0.415124in}{0.220679in}}{\pgfqpoint{1.785856in}{1.579782in}}%
\pgfusepath{clip}%
\pgfsetroundcap%
\pgfsetroundjoin%
\pgfsetlinewidth{0.803000pt}%
\definecolor{currentstroke}{rgb}{0.800000,0.800000,0.800000}%
\pgfsetstrokecolor{currentstroke}%
\pgfsetdash{}{0pt}%
\pgfpathmoveto{\pgfqpoint{0.415124in}{1.265374in}}%
\pgfpathlineto{\pgfqpoint{2.200980in}{1.265374in}}%
\pgfusepath{stroke}%
\end{pgfscope}%
\begin{pgfscope}%
\definecolor{textcolor}{rgb}{0.150000,0.150000,0.150000}%
\pgfsetstrokecolor{textcolor}%
\pgfsetfillcolor{textcolor}%
\pgftext[x=0.179012in, y=1.217148in, left, base]{\color{textcolor}\rmfamily\fontsize{10.000000}{12.000000}\selectfont \(\displaystyle {20}\)}%
\end{pgfscope}%
\begin{pgfscope}%
\pgfpathrectangle{\pgfqpoint{0.415124in}{0.220679in}}{\pgfqpoint{1.785856in}{1.579782in}}%
\pgfusepath{clip}%
\pgfsetroundcap%
\pgfsetroundjoin%
\pgfsetlinewidth{0.803000pt}%
\definecolor{currentstroke}{rgb}{0.800000,0.800000,0.800000}%
\pgfsetstrokecolor{currentstroke}%
\pgfsetdash{}{0pt}%
\pgfpathmoveto{\pgfqpoint{0.415124in}{1.728653in}}%
\pgfpathlineto{\pgfqpoint{2.200980in}{1.728653in}}%
\pgfusepath{stroke}%
\end{pgfscope}%
\begin{pgfscope}%
\definecolor{textcolor}{rgb}{0.150000,0.150000,0.150000}%
\pgfsetstrokecolor{textcolor}%
\pgfsetfillcolor{textcolor}%
\pgftext[x=0.179012in, y=1.680428in, left, base]{\color{textcolor}\rmfamily\fontsize{10.000000}{12.000000}\selectfont \(\displaystyle {30}\)}%
\end{pgfscope}%
\begin{pgfscope}%
\definecolor{textcolor}{rgb}{0.150000,0.150000,0.150000}%
\pgfsetstrokecolor{textcolor}%
\pgfsetfillcolor{textcolor}%
\pgftext[x=0.123457in,y=1.010570in,,bottom,rotate=90.000000]{\color{textcolor}\rmfamily\fontsize{10.000000}{12.000000}\selectfont Nubmer of terms}%
\end{pgfscope}%
\begin{pgfscope}%
\pgfpathrectangle{\pgfqpoint{0.415124in}{0.220679in}}{\pgfqpoint{1.785856in}{1.579782in}}%
\pgfusepath{clip}%
\pgfsetbuttcap%
\pgfsetmiterjoin%
\definecolor{currentfill}{rgb}{0.827451,0.827451,0.827451}%
\pgfsetfillcolor{currentfill}%
\pgfsetlinewidth{1.003750pt}%
\definecolor{currentstroke}{rgb}{0.498039,0.498039,0.498039}%
\pgfsetstrokecolor{currentstroke}%
\pgfsetdash{}{0pt}%
\pgfpathmoveto{\pgfqpoint{0.504417in}{0.616783in}}%
\pgfpathlineto{\pgfqpoint{1.218759in}{0.616783in}}%
\pgfpathlineto{\pgfqpoint{1.218759in}{0.765233in}}%
\pgfpathlineto{\pgfqpoint{1.040173in}{0.802094in}}%
\pgfpathlineto{\pgfqpoint{1.218759in}{0.838956in}}%
\pgfpathlineto{\pgfqpoint{1.218759in}{1.219046in}}%
\pgfpathlineto{\pgfqpoint{0.504417in}{1.219046in}}%
\pgfpathlineto{\pgfqpoint{0.504417in}{0.838956in}}%
\pgfpathlineto{\pgfqpoint{0.683002in}{0.802094in}}%
\pgfpathlineto{\pgfqpoint{0.504417in}{0.765233in}}%
\pgfpathlineto{\pgfqpoint{0.504417in}{0.616783in}}%
\pgfpathlineto{\pgfqpoint{0.504417in}{0.616783in}}%
\pgfpathclose%
\pgfusepath{stroke,fill}%
\end{pgfscope}%
\begin{pgfscope}%
\pgfpathrectangle{\pgfqpoint{0.415124in}{0.220679in}}{\pgfqpoint{1.785856in}{1.579782in}}%
\pgfusepath{clip}%
\pgfsetbuttcap%
\pgfsetmiterjoin%
\definecolor{currentfill}{rgb}{0.827451,0.827451,0.827451}%
\pgfsetfillcolor{currentfill}%
\pgfsetlinewidth{1.003750pt}%
\definecolor{currentstroke}{rgb}{0.498039,0.498039,0.498039}%
\pgfsetstrokecolor{currentstroke}%
\pgfsetdash{}{0pt}%
\pgfpathmoveto{\pgfqpoint{1.397345in}{0.709439in}}%
\pgfpathlineto{\pgfqpoint{2.111687in}{0.709439in}}%
\pgfpathlineto{\pgfqpoint{2.111687in}{0.768042in}}%
\pgfpathlineto{\pgfqpoint{1.933102in}{0.802094in}}%
\pgfpathlineto{\pgfqpoint{2.111687in}{0.836146in}}%
\pgfpathlineto{\pgfqpoint{2.111687in}{1.265374in}}%
\pgfpathlineto{\pgfqpoint{1.397345in}{1.265374in}}%
\pgfpathlineto{\pgfqpoint{1.397345in}{0.836146in}}%
\pgfpathlineto{\pgfqpoint{1.575930in}{0.802094in}}%
\pgfpathlineto{\pgfqpoint{1.397345in}{0.768042in}}%
\pgfpathlineto{\pgfqpoint{1.397345in}{0.709439in}}%
\pgfpathlineto{\pgfqpoint{1.397345in}{0.709439in}}%
\pgfpathclose%
\pgfusepath{stroke,fill}%
\end{pgfscope}%
\begin{pgfscope}%
\pgfpathrectangle{\pgfqpoint{0.415124in}{0.220679in}}{\pgfqpoint{1.785856in}{1.579782in}}%
\pgfusepath{clip}%
\pgfsetroundcap%
\pgfsetroundjoin%
\pgfsetlinewidth{1.003750pt}%
\definecolor{currentstroke}{rgb}{0.498039,0.498039,0.498039}%
\pgfsetstrokecolor{currentstroke}%
\pgfsetdash{}{0pt}%
\pgfpathmoveto{\pgfqpoint{0.861588in}{0.616783in}}%
\pgfpathlineto{\pgfqpoint{0.861588in}{0.477799in}}%
\pgfusepath{stroke}%
\end{pgfscope}%
\begin{pgfscope}%
\pgfpathrectangle{\pgfqpoint{0.415124in}{0.220679in}}{\pgfqpoint{1.785856in}{1.579782in}}%
\pgfusepath{clip}%
\pgfsetroundcap%
\pgfsetroundjoin%
\pgfsetlinewidth{1.003750pt}%
\definecolor{currentstroke}{rgb}{0.498039,0.498039,0.498039}%
\pgfsetstrokecolor{currentstroke}%
\pgfsetdash{}{0pt}%
\pgfpathmoveto{\pgfqpoint{0.861588in}{1.219046in}}%
\pgfpathlineto{\pgfqpoint{0.861588in}{1.728653in}}%
\pgfusepath{stroke}%
\end{pgfscope}%
\begin{pgfscope}%
\pgfpathrectangle{\pgfqpoint{0.415124in}{0.220679in}}{\pgfqpoint{1.785856in}{1.579782in}}%
\pgfusepath{clip}%
\pgfsetroundcap%
\pgfsetroundjoin%
\pgfsetlinewidth{1.003750pt}%
\definecolor{currentstroke}{rgb}{0.498039,0.498039,0.498039}%
\pgfsetstrokecolor{currentstroke}%
\pgfsetdash{}{0pt}%
\pgfpathmoveto{\pgfqpoint{0.683002in}{0.477799in}}%
\pgfpathlineto{\pgfqpoint{1.040173in}{0.477799in}}%
\pgfusepath{stroke}%
\end{pgfscope}%
\begin{pgfscope}%
\pgfpathrectangle{\pgfqpoint{0.415124in}{0.220679in}}{\pgfqpoint{1.785856in}{1.579782in}}%
\pgfusepath{clip}%
\pgfsetroundcap%
\pgfsetroundjoin%
\pgfsetlinewidth{1.003750pt}%
\definecolor{currentstroke}{rgb}{0.498039,0.498039,0.498039}%
\pgfsetstrokecolor{currentstroke}%
\pgfsetdash{}{0pt}%
\pgfpathmoveto{\pgfqpoint{0.683002in}{1.728653in}}%
\pgfpathlineto{\pgfqpoint{1.040173in}{1.728653in}}%
\pgfusepath{stroke}%
\end{pgfscope}%
\begin{pgfscope}%
\pgfpathrectangle{\pgfqpoint{0.415124in}{0.220679in}}{\pgfqpoint{1.785856in}{1.579782in}}%
\pgfusepath{clip}%
\pgfsetroundcap%
\pgfsetroundjoin%
\pgfsetlinewidth{1.003750pt}%
\definecolor{currentstroke}{rgb}{0.498039,0.498039,0.498039}%
\pgfsetstrokecolor{currentstroke}%
\pgfsetdash{}{0pt}%
\pgfpathmoveto{\pgfqpoint{1.754516in}{0.709439in}}%
\pgfpathlineto{\pgfqpoint{1.754516in}{0.292487in}}%
\pgfusepath{stroke}%
\end{pgfscope}%
\begin{pgfscope}%
\pgfpathrectangle{\pgfqpoint{0.415124in}{0.220679in}}{\pgfqpoint{1.785856in}{1.579782in}}%
\pgfusepath{clip}%
\pgfsetroundcap%
\pgfsetroundjoin%
\pgfsetlinewidth{1.003750pt}%
\definecolor{currentstroke}{rgb}{0.498039,0.498039,0.498039}%
\pgfsetstrokecolor{currentstroke}%
\pgfsetdash{}{0pt}%
\pgfpathmoveto{\pgfqpoint{1.754516in}{1.265374in}}%
\pgfpathlineto{\pgfqpoint{1.754516in}{1.728653in}}%
\pgfusepath{stroke}%
\end{pgfscope}%
\begin{pgfscope}%
\pgfpathrectangle{\pgfqpoint{0.415124in}{0.220679in}}{\pgfqpoint{1.785856in}{1.579782in}}%
\pgfusepath{clip}%
\pgfsetroundcap%
\pgfsetroundjoin%
\pgfsetlinewidth{1.003750pt}%
\definecolor{currentstroke}{rgb}{0.498039,0.498039,0.498039}%
\pgfsetstrokecolor{currentstroke}%
\pgfsetdash{}{0pt}%
\pgfpathmoveto{\pgfqpoint{1.575930in}{0.292487in}}%
\pgfpathlineto{\pgfqpoint{1.933102in}{0.292487in}}%
\pgfusepath{stroke}%
\end{pgfscope}%
\begin{pgfscope}%
\pgfpathrectangle{\pgfqpoint{0.415124in}{0.220679in}}{\pgfqpoint{1.785856in}{1.579782in}}%
\pgfusepath{clip}%
\pgfsetroundcap%
\pgfsetroundjoin%
\pgfsetlinewidth{1.003750pt}%
\definecolor{currentstroke}{rgb}{0.498039,0.498039,0.498039}%
\pgfsetstrokecolor{currentstroke}%
\pgfsetdash{}{0pt}%
\pgfpathmoveto{\pgfqpoint{1.575930in}{1.728653in}}%
\pgfpathlineto{\pgfqpoint{1.933102in}{1.728653in}}%
\pgfusepath{stroke}%
\end{pgfscope}%
\begin{pgfscope}%
\pgfpathrectangle{\pgfqpoint{0.415124in}{0.220679in}}{\pgfqpoint{1.785856in}{1.579782in}}%
\pgfusepath{clip}%
\pgfsetroundcap%
\pgfsetroundjoin%
\pgfsetlinewidth{1.003750pt}%
\definecolor{currentstroke}{rgb}{0.498039,0.498039,0.498039}%
\pgfsetstrokecolor{currentstroke}%
\pgfsetdash{}{0pt}%
\pgfpathmoveto{\pgfqpoint{0.683002in}{0.802094in}}%
\pgfpathlineto{\pgfqpoint{1.040173in}{0.802094in}}%
\pgfusepath{stroke}%
\end{pgfscope}%
\begin{pgfscope}%
\pgfpathrectangle{\pgfqpoint{0.415124in}{0.220679in}}{\pgfqpoint{1.785856in}{1.579782in}}%
\pgfusepath{clip}%
\pgfsetroundcap%
\pgfsetroundjoin%
\pgfsetlinewidth{1.003750pt}%
\definecolor{currentstroke}{rgb}{0.498039,0.498039,0.498039}%
\pgfsetstrokecolor{currentstroke}%
\pgfsetdash{}{0pt}%
\pgfpathmoveto{\pgfqpoint{1.575930in}{0.802094in}}%
\pgfpathlineto{\pgfqpoint{1.933102in}{0.802094in}}%
\pgfusepath{stroke}%
\end{pgfscope}%
\begin{pgfscope}%
\pgfsetrectcap%
\pgfsetmiterjoin%
\pgfsetlinewidth{0.803000pt}%
\definecolor{currentstroke}{rgb}{0.800000,0.800000,0.800000}%
\pgfsetstrokecolor{currentstroke}%
\pgfsetdash{}{0pt}%
\pgfpathmoveto{\pgfqpoint{0.415124in}{0.220679in}}%
\pgfpathlineto{\pgfqpoint{0.415124in}{1.800461in}}%
\pgfusepath{stroke}%
\end{pgfscope}%
\begin{pgfscope}%
\pgfsetrectcap%
\pgfsetmiterjoin%
\pgfsetlinewidth{0.803000pt}%
\definecolor{currentstroke}{rgb}{0.800000,0.800000,0.800000}%
\pgfsetstrokecolor{currentstroke}%
\pgfsetdash{}{0pt}%
\pgfpathmoveto{\pgfqpoint{2.200980in}{0.220679in}}%
\pgfpathlineto{\pgfqpoint{2.200980in}{1.800461in}}%
\pgfusepath{stroke}%
\end{pgfscope}%
\begin{pgfscope}%
\pgfsetrectcap%
\pgfsetmiterjoin%
\pgfsetlinewidth{0.803000pt}%
\definecolor{currentstroke}{rgb}{0.800000,0.800000,0.800000}%
\pgfsetstrokecolor{currentstroke}%
\pgfsetdash{}{0pt}%
\pgfpathmoveto{\pgfqpoint{0.415124in}{0.220679in}}%
\pgfpathlineto{\pgfqpoint{2.200980in}{0.220679in}}%
\pgfusepath{stroke}%
\end{pgfscope}%
\begin{pgfscope}%
\pgfsetrectcap%
\pgfsetmiterjoin%
\pgfsetlinewidth{0.803000pt}%
\definecolor{currentstroke}{rgb}{0.800000,0.800000,0.800000}%
\pgfsetstrokecolor{currentstroke}%
\pgfsetdash{}{0pt}%
\pgfpathmoveto{\pgfqpoint{0.415124in}{1.800461in}}%
\pgfpathlineto{\pgfqpoint{2.200980in}{1.800461in}}%
\pgfusepath{stroke}%
\end{pgfscope}%
\end{pgfpicture}%
\makeatother%
\endgroup%